\documentclass[10pt,twocolumn,letterpaper]{article}

\usepackage[pagenumbers]{cvpr} 

\usepackage{graphicx}
\usepackage{amsmath}
\usepackage{amssymb}
\usepackage{booktabs}
\usepackage{sidecap}
\usepackage[accsupp]{axessibility}
\usepackage{array}

\usepackage[pagebackref,breaklinks,colorlinks]{hyperref}

\usepackage[capitalize]{cleveref}
\crefname{section}{Sec.}{Secs.}
\Crefname{section}{Section}{Sections}
\Crefname{table}{Table}{Tables}
\crefname{table}{Tab.}{Tabs.}

\def\cvprPaperID{8975} 
\def\confName{CVPR}
\def\confYear{2023}

\begin{document}

\title{3D-POP - An automated annotation approach to facilitate markerless 2D-3D tracking of freely moving birds
with marker-based motion capture}
\author{
Hemal Naik$^{1234*}$,Alex Hoi Hang Chan$^{12*}$,Junran Yang$^{2}$, Mathilde Delacoux$^{12}$,\\Iain D. Couzin$^{123}$, Fumihiro Kano$^{12\dag}$, Máté Nagy$^{12356\dag}$ \\ \\
\normalsize $^1$Dept. of Collective Behavior and Dept. of Ecology of Animal Societies, Max Planck Institute of Animal Behavior, \\
\normalsize $^2$Dept. of Biology, University of Konstanz, $^3$Centre for the Advanced Study of Collective Behaviour,  University of Konstanz, \\
\normalsize$^4$Computer Aided Medial Procedures, Informatik Department, Technische Universität München,\\ 
\normalsize $^5$Dept. of Biological Physics, Eötvös Loránd University, $^6$MTA-ELTE ‘Lendület’ Collective Behaviour Research Group,\\
\normalsize  Hungarian Academy of Sciences.
}

\maketitle
\begin{abstract}
Recent advances in machine learning and computer vision are revolutionizing the field of animal behavior by enabling researchers to track the poses and locations of freely moving animals without any marker attachment. 
However, large datasets of annotated images of animals for markerless pose tracking, especially high-resolution images taken from multiple angles with accurate 3D annotations, are still scant. 
Here, we propose a method that uses a motion capture (mo-cap) system to obtain a large amount of annotated data on animal movement and posture (2D and 3D) in a semi-automatic manner. 
Our method is novel in that it extracts the 3D positions of morphological keypoints (e.g eyes, beak, tail) in reference to the positions of markers attached to the animals. 
Using this method, we obtained, and offer here, a new dataset - 3D-POP with approximately 300k annotated frames (4 million instances) in the form of videos having groups of one to ten freely moving birds from 4 different camera views in a 3.6m x 4.2m area. 
3D-POP is the first dataset of flocking birds with accurate keypoint annotations in 2D and 3D along with bounding box and individual identities and will facilitate the development of solutions for problems of 2D to 3D markerless pose, trajectory tracking, and identification in birds.
\end{abstract}

\begin{figure}
\centering
   \includegraphics[width=0.7\linewidth,trim=0cm 1.5cm 0cm 0cm]{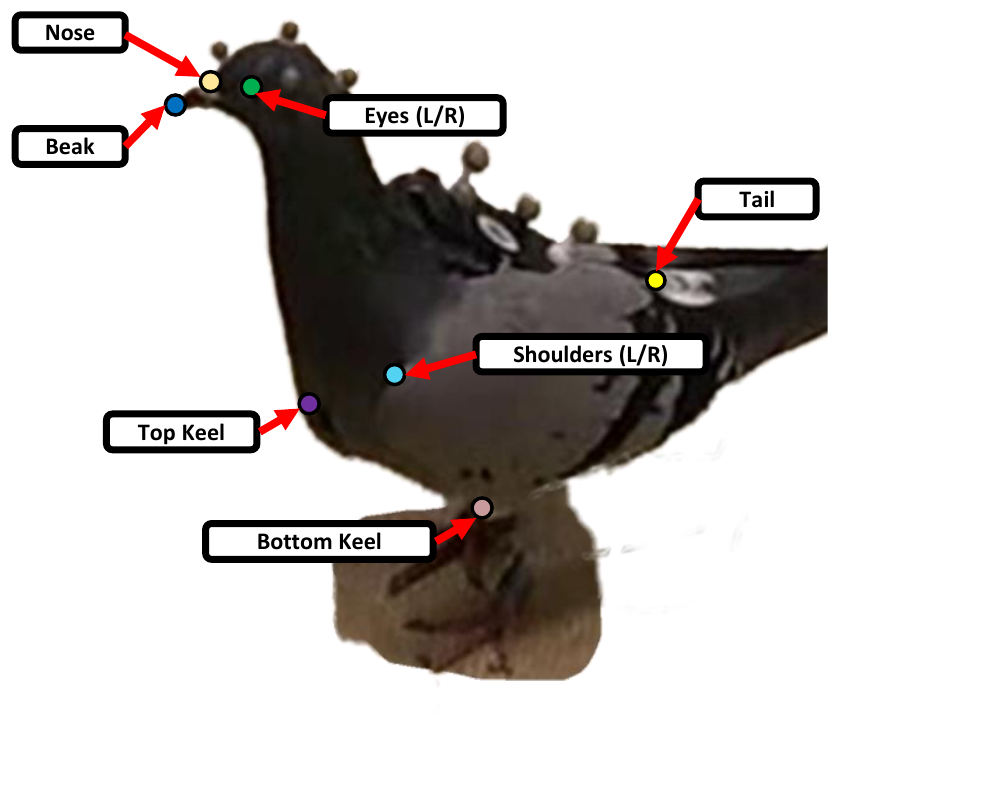}
   \caption{Definition of morphological keypoints offered in the 3D-POP dataset }
\label{fig:keypoints}
\end{figure}

\section{Introduction}
Computer vision and machine learning are revolutionizing many facets of conventional research methods. 
For example, dataset-driven machine learning methods have demonstrated remarkable success in the field of animal behavior, in tasks related to object detection \cite{van2018inaturalist,wah_caltech-ucsd_2011}, tracking and individual identification \cite{walter2021trex,ferreira2020deep,lauer_multi-animal_2022}, species recognition \cite{van2018inaturalist}, 2D pose estimation \cite{mathis_deep_2020,graving_deepposekit_2019} and 3D pose estimation \cite{badger_3d_2020,biggs_creatures_2019}. 
These automatic methods not only reduce the required labor and errors associated with manual coding of behaviours \cite{tuia_perspectives_2022,borowiec_deep_2022} but also facilitate long-term continuous monitoring of animal behavior in both indoor (lab) \cite{stowers2017virtual,nourizonoz2020etholoop} and outdoor (wild) settings \cite{swanson2015snapshot,gagne2021florida}. 
Engineering and robotics experts use the data on animal locomotion to reverse-engineer the key mechanisms underlying behaviors and movements of animals \cite{joska2021acinoset,jafferis2019untethered}. 
The development of new techniques critically depends on the quality of publicly-available datasets with accurate annotations. 

Creating large datasets with animals is particularly difficult because every species has distinct morphology, and also because it is generally challenging to film freely moving animals in a controlled environment. 
It is thus important for datasets to include a wide range of species and behaviors to maximize the practical application of machine learning methods for animal tracking. 
Although animals have been included in many popular image datasets collected from the internet such as ImageNet \cite{deng_imagenet_2009} and COCO \cite{lin_microsoft_2014}, those datasets have not fulfilled more specific needs of animal behavior researchers.  
Hence, recently several datasets have been created with a focus on animal behavior research, such as species classification  \cite{swanson2015snapshot,gagne2021florida,van2018inaturalist,Yu_2021,wah_caltech-ucsd_2011}, behavioral classification \cite{ng_animal_2022,Yu_2021,marshall_pair-r24m_2021} and posture tracking \cite{yao_openmonkeychallenge_2022,gunel2019deepfly3d,badger_3d_2020,dunn2021geometric,labuguen2021macaquepose}.

The most common approach for creating datasets of animals is through manual annotations in the image space (2D).
As a result, most solutions to single / multiple animal detection, tracking, or pose estimation problems are limited to the 2D space \cite{lauer_multi-animal_2022, graving_deepposekit_2019}, or use 2D image projections to validate the results of 3D predictions without ground truth \cite{biggs_creatures_2019,badger_3d_2020}.
For nonhuman animals, a dataset similar to Human 3.6M \cite{h36m_pami} is necessary to develop solutions for problems on 2D/3D tracking and posture prediction with a range of constraints, such as single or multiviews, single or multi-individual, and tracking using single frame or temporal consistency. 
More recently, marker-based motion-capture technology has been used to create 3D datasets for rats \cite{dunn2021geometric} and dogs \cite{kearney2020rgbd} with one individual. 
The application of mo-cap for animal behaviour studies has also increased in popularity, such as studying flight kinematics \cite{kleinheerenbrink2022optimization} and gaze behavior in a freely moving group \cite{kano2022birds, itahara2022corvid}.
It is clear that datasets with mo-cap will not only enhance the size of the dataset but also improve the accuracy of annotations, thus providing a large 2D/3D ground truth dataset for the animal position, posture, and identity tracking. 
However, despite its potential, researchers have only begun using mo-cap for behavior studies, and further work is required in terms of method development and dataset collection.

We propose a new mo-cap-based approach to create large-scale datasets with a bird species (homing pigeons, \textit{Columba livia}), and provide a complete code base for further applications to other species.
Along with 2D-3D posture, the dataset also offers annotations for 2D-3D movement trajectories (position) with ground truth on identities for up to 18 individuals. 
We overcame the unique challenge of needing to attach reflective markers on desired but often inaccessible morphological keypoints on animal bodies and instead determined the relative 3D position of these keypoints to markers attached on accessible parts of the animal (Figure \ref{fig:keypoints}).

The method enables a large amount of training data to be generated in a semi-automatic manner with minimal time investment and human labor. 
Moreover, by tracking freely-moving animals in a relatively large area (3.6m x 4.2m), we were able to track a variety of naturalistic behaviors in a flock consisting of up to 10 individuals under realistic experimental conditions.
Finally, we demonstrate through a series of experiments that our method is consistent and the CNN models trained on our dataset are able to predict the postures of birds with no markers attached to their bodies.

\section{State of the Art}

\begin{SCfigure*}[\sidecaptionrelwidth]
\centering 
   \includegraphics[width=0.8\textwidth]{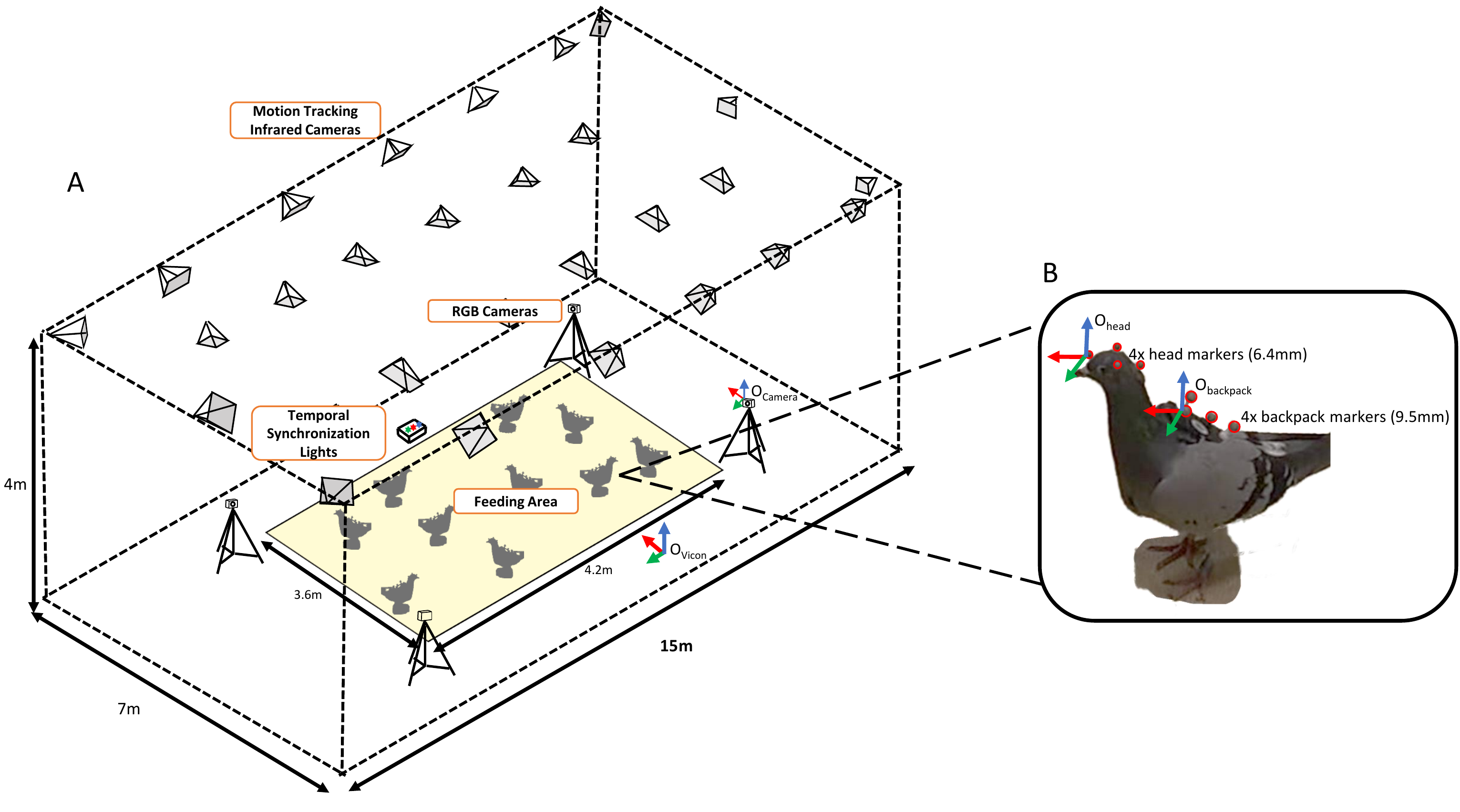}
   \caption{Illustration showing the experimental setup, with different defined coordinate systems, including the Vicon global coordinate system (O$_{\text{Vicon}}$), the camera coordinate system for each RGB camera (O$_{\text{Camera}}$), and the head  (O$_{\text{head}}$) and backpack  (O$_{\text{backpack}}$) coordinate system for each pigeon subject  A) Detailed floor plan for data collection. B) Pigeon subject, with corresponding head and backpack markers and coordinate systems}
\label{fig:ExpSetup}
\end{SCfigure*}

\subsection{2D posture}
Animal Kingdom \cite{ng_animal_2022} is by far the largest dataset with 50 hours of video annotations that include 850 species of varied taxa (fish, birds, mammals, etc.), focusing on a generalizable solution for 2D pose estimation and activity recognition for a single individual. Other notable datasets contain images instead of videos and focus on capturing variations in terms of specific taxa \eg mammals \cite{Yu_2021}, birds \cite{wah_caltech-ucsd_2011} and monkeys \cite{yao_openmonkeychallenge_2022}, or specific species \eg zebras \cite{graving_deepposekit_2019}, all of these focus on solving problems for a single individual recorded from a single viewpoint.

Datasets based on single animal-based solutions are sufficient for some cases and rely on detection-based top-down approaches for extending the method for tracking the posture of multiple individuals \cite{waldmann2022muppet}.
There are few datasets that offer posture annotations for multiple individuals \cite{lauer_multi-animal_2022, labuguen2021macaquepose, badger_3d_2020}.
The problem of tracking multiple individuals is often simplified by placing the cameras above the animals, which minimizes occlusions \cite{walter2021trex,lauer_multi-animal_2022}.
Tracking multiple individuals from side views may require multiple views, which may be important to resolve occlusions when animals interact in 3D spaces \eg Cowbird dataset \cite{badger_3d_2020}.

The existing datasets have motivated the development of various methods for posture estimation.
However, reliance on manual annotations limits the complexity of datasets in terms of the number of viewpoints or the number of individuals, especially for video sequences.

\subsection{3D posture} \label{ss:3DPosture}
Datasets with ground truth on 3D posture are relatively difficult to obtain with a group of animals.
One popular method for obtaining 3D ground truth posture is the triangulation of 2D postures using multiple views to record animals. 
Acinoset \cite{joska2021acinoset} (leopard in wild), Fly3D \cite{gunel2019deepfly3d} (fly in a lab) and OpenMonkeyStudio \cite{bala2020automated} (macaque in a lab) use triangulation-based approaches to provide 3D posture of single individuals.
The images for these datasets are also annotated manually and, therefore, the accuracy of the computed 3D pose depends on the quality of annotation and calibration.

An alternative approach is to use marker-based mo-cap with a skeleton tracking feature as used with humans \cite{h36m_pami}.
Kearney \etal \cite{kearney2020rgbd} used motion capture to generate 3D ground truth for dogs and combine their approach with depth sensors (RGB-D) with the aim of designing markerless tracking based on RGBD sensors (63 to 82 markers).
Dunn \etal \cite{dunn2021geometric} offered Rat 7M dataset using mo-cap with RGB cameras and 20 markers.   
These datasets are useful for solving posture problems for a single individual from multiple views and offer the option of using temporal consistency.

Recently, Marshall \etal published PAIR-R24M \cite{marshall_pair-r24m_2021}, the first dataset with 3D ground truth with more than one animal, a pair of rats, using the approach of Dunn \etal \cite{dunn2021geometric}.
Motion capture systems offer the huge advantage of creating millions of annotations in an automatic manner with high accuracy and low noise.
The skeleton tracking feature with the mo-cap is primarily designed for tracking human posture and relies on a large number of markers. 
Additionally, the marker patterns have to be unique (at least partially) for maintaining the identity of each individual. 
Marker placement is a also limitation for smaller species and wild animals that lack the tolerance for having markers placed on specific locations of their body.

Lack of ground truth in 3D posture had led to innovative work of predicting 3D posture using 2D keypoints and silhouettes \cite{badger_3d_2020,biggs_creatures_2019} or using synthetic datasets \cite{bolanos2021three} or toys \cite{zuffi20173d}.
These approaches are promising but lack quantitative evaluation for robust practical applications. 
Computer vision literature on 3D posture problems mainly focuses on extracting as much of detail as possible.
For animal behavior experiments information required from videos is always defined in the context of the experiment.
It is worth noting that for birds, tracking the head and body orientations could be sufficient to quantify many key behaviors in ground-foraging contexts, such as feeding (pecking ground), preening, vigilance (head scanning), courtship (head bowing), or walking. 
Measuring the head direction in 3D also allows gaze reconstruction \cite{kano2022birds,itahara2022corvid}, to be applied in the study of social cognition and collective behavior.

\subsection{Multi-object tracking with identity}
Identity recognition is a critical problem to solve in the context of biological studies, especially when tracking the behavior of multiple interacting individuals over long periods of time. Tracking and identification of multiple individuals in large groups are especially exciting to quantify group-level behaviors like social networks \cite{whitehead1997analysing,xiao2022multi}, dominance, or leadership \cite{nagy2013context}. 

For indoor experiments, the number of individuals is often controlled and the identification problem is linked with the tracking of animals \cite{walter2021trex,lauer_multi-animal_2022}.
The task of tracking and identification is often resolved together using markers \cite{graving_deepposekit_2019, nagy2013context} or marker-less \cite{walter2021trex, bozek2021markerless, ferreira2020deep} methods. 
The existing solutions perform well with specific perspectives (top-down view) and thus often fail to resolve cases of occlusion.
Robust evaluation of simultaneous identification and tracking methods is difficult because true ground truth for identities is often not available in datasets with multiple animals or available for only a very short duration \cite{lauer_multi-animal_2022}.

There are many good datasets available to independently solve problems of posture estimation, detection, tracking, and identification.
Very few datasets offer the possibility of solving all of these problems simultaneously in realistic experimental scenarios.

We aim to fill this gap with our contribution of a semi-automatic method for producing new datasets with animals.
Our dataset, 3D-POP, includes video recordings of 18 unique pigeons in various group sizes (1,2,5,10) from multiple views.
We offer ground truth for identity, 2D-3D trajectories, and 2D-3D posture mapping for all individuals across the entire dataset (300K frames).
The dataset also consists of annotations for object detection in the form of bounding boxes. 

\section{Methods}
\subsection{Experimental Setup}
The dataset was collected from pigeons moving on a jute fabric (3.6m x 4.2m) onto which we evenly scattered grains to encourage the birds to feed in that area (Figure \ref{fig:ExpSetup}A). 
This feeding area was located inside a large enclosure equipped with a mo-cap system (15m x 7m x 4m).
This mo-cap system consists of 30 motion capture cameras (12 Vicon Vero 2.2, 18 Vicon Vantage-5 cameras; 100Hz) and can track the 3D positions of reflective markers with sub-millimeter precision. 
At the corners of the feeding area, we placed 4 high-resolution (4K) Sony action cameras (rx0ii, 30Hz, 3840x2160p) mounted on standard tripods and also an Arduino-based synchronization box which flashes RGB and infrared LED lights every 5 seconds (Figure \ref{fig:ExpSetup}). 
Details on the synchronization and calibration of RGB cameras are provided in the supplementary text.

\subsection{Animal Subjects}
Eighteen pigeons (\textit{Columba livia}) were subjected to this study over 6 experimental days. 
Each day 10 pigeons were randomly selected from the population. 
Four 6.4mm reflective markers were attached to each subject's head, and four 9.5mm markers were attached to a customized backpack worn by each subject (Figure \ref{fig:ExpSetup}B). 
Generally, pigeons tolerate markers on the head with minimal effects on their behavior and habituate quickly to backpacks.
Backpacks are also widely used for bird studies in behavioral ecology \cite{alarcon2018automated,williamson2021lightweight}. 
The four 9.5mm backpack markers had a unique geometric configuration to track the individual identities of each bird throughout each recording.
Each day we performed up to 11 trials in the following order: 1 pigeon (4 trials), a pair of pigeons (4 trials), a flock of 5 pigeons (2 trials), and a flock of 10 pigeons (1 trial). 
It took approximately 1 hour to perform all trials each day. 
The total frames and duration of samples over the course of the experiment are described in Table \ref{table:datasum}. 
An additional session was recorded with birds without attaching any markers to validate the results of models trained on annotated data having birds with markers (see \ref{ss:Exp2}).

\subsection{Data annotation pipeline}

\begin{SCfigure*}[\sidecaptionrelwidth][t]
\centering 
   \includegraphics[width=0.7\textwidth, trim = 3cm 1cm 3cm 0.5cm]{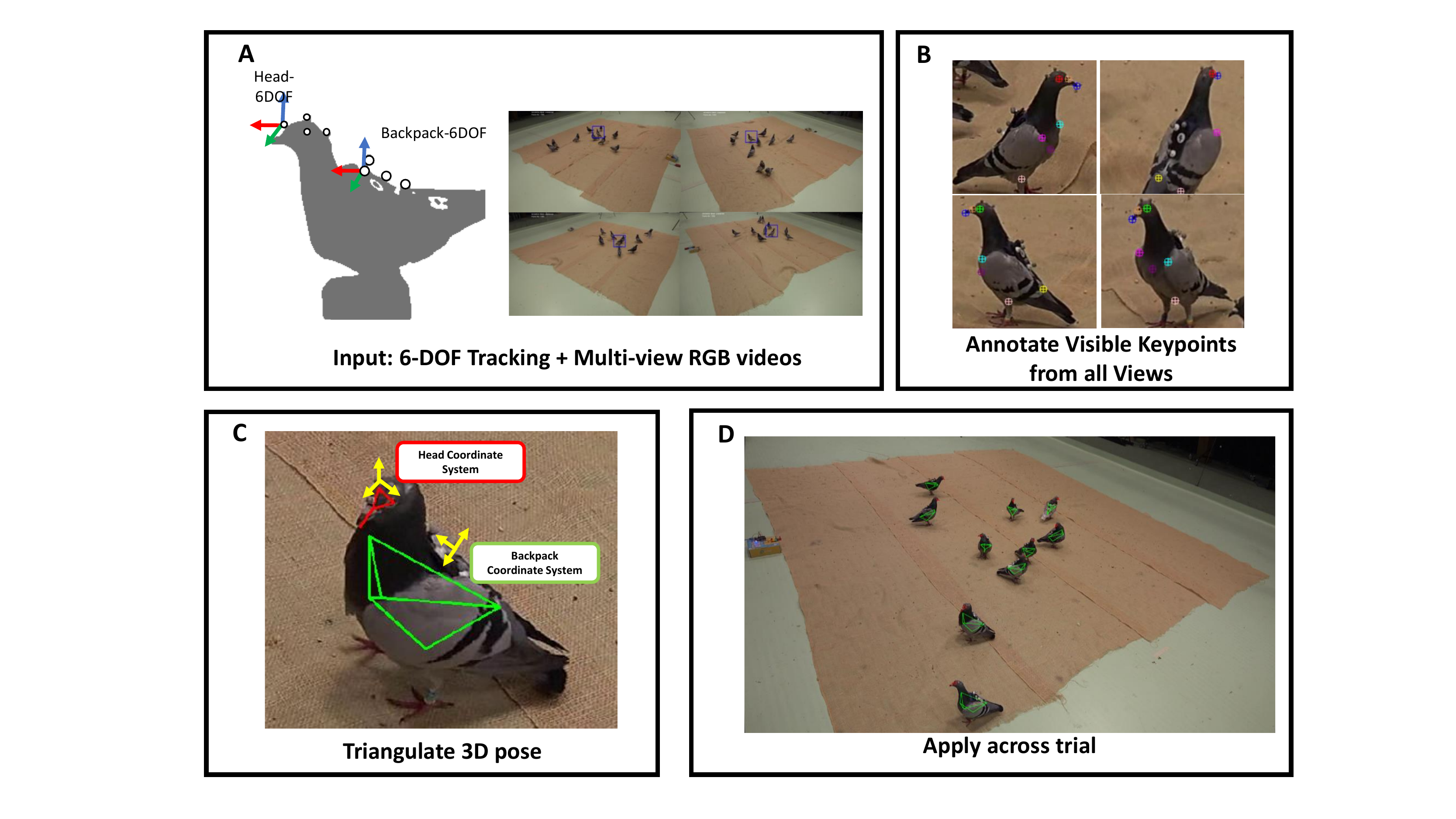}
   \caption{Semi-automated annotation pipeline based on 6DOF tracking and RGB images. A) Input 6-DOF tracking data for head and backpack coordinate systems, and multi-view RGB videos. B) Manually annotate all visible keypoints from all views. C) Triangulate 3D position of all keypoints in the head and backpack coordinate system, assuming that keypoints and tracked markers are a rigid body. D) Apply across trials to get keypoints across all individuals}
\label{fig:AnnotaPipeline}
\end{SCfigure*}

\subsubsection{Annotation principle}
The movement of all features on a rigid body can be tracked simultaneously in a 3D space by computing 6-DOF pose of the rigid object. 
We use this principle to achieve annotations for keypoint features that are rigidly attached to the head and body of the bird. 
The four markers attached to the head and body (using a backpack) of each pigeon are used to compute 6-DOF pose of these body parts using the mo-cap system.

By assuming that the head and body are rigid bodies in the case of walking or standing birds, we designed a pipeline to annotate the position of features on the head and body (beak, eyes, shoulder, and tail, etc.) in a few frames to compute their 3D location with respect to marker positions.
Once computed, the relationship between markers and features does not change during the sequences and this ensures that 6-DOF pose of head and body for any frame can be used to project 3D positions of keypoint features onto the image space to obtain 2D annotations.

All keypoints defined for the head lie on the skull of the bird (Figure \ref{fig:keypoints}).
The rigidity assumption is valid for these keypoints as they are rigidly placed on the skull. The keypoints chosen for the body lie actually on the rib cage and shoulders and exhibit a limited range of motion independent of each other. 
The rigidity assumption for the body is a reasonable assumption for the annotation pipeline if the birds do not move their wings and body(see \ref{ss:Exp3}).

\subsubsection{Manual annotation}
6-DOF (Degrees of freedom) tracking of the head and body is used to create a bounding box around the bird and crop the image of the focal individual for annotation.
For each individual pigeon, 9 morphological keypoints (Figure \ref{fig:keypoints}) are annotated on 5-10 frames from all available view angles. 
Ideally, four frames (1 per view) is sufficient, but all keypoints are rarely visible within a single instance.
Moreover, multiple measurements (3-5 frames per view) improve the robustness of computed 3D keypoint positions.
The position of each keypoint is first triangulated using sparse bundle adjustment (in the camera coordinate system), then
the relative position of the keypoint is computed with respect to the markers (in the coordinate system of the body part). 
Finally, all resultant 3D positions of keypoints are averaged and stored as a template file.
This process is repeated for each bird on each recording day.

\subsubsection{Annotation propagation}
In this final step, the ground truth data is generated for each recording using 3D keypoint positions computed in the previous step. 
The 3D positions of the keypoint features are transferred to the global coordinate system using 6-DOF pose.
Next, keypoints are transferred to the coordinate system of each camera and projected to the image space (using calibration parameters).
Bounding box annotations for object detection or tracking tasks are derived from keypoint projections. 
We determined that keypoints with the minimum and maximum x-y pixel values with an offset of 60 pixels are sufficient to define a bounding box.
Finally, the 6-DOF tracking with the mo-cap system maintains the identity of each bird and this is also stored with 2D-3D information for the entire sequence. 

\section{The 3D-POP dataset}
\subsection{Dataset Description}
We present 3D-POP (3D Postures of Pigeons), a dataset that provides accurate ground truth for 3D keypoints, 2D keypoints, bounding boxes, and individual identities.
The dataset includes RGB images from four high-resolution cameras (4K) and up to 6 hours of recordings divided into 57 sequences of 1,2,5, and 10 pigeons behaving naturalistically (Table \ref{table:datasum}).
The dataset contains 3D coordinates (unfiltered) and 6-DOF pose obtained from the mo-cap facility along with calibration parameters. 
We also provide a total of 1 hour of recording (11 sequences) with pigeons (group size:1,2,5,11) without any markers on their body. 
These videos are provided for users to test the practical effectiveness of markerless solutions without the influence of markers.
For realistic assessment, we show that a model trained with our dataset is able to infer keypoints on videos with pigeons without markers (see \ref{ss:Exp2}).
Download the dataset here: https://tinyurl.com/4ckbjcpx

\begin{table}[b]
\begin{center}
\newcolumntype{C}[1]{>{\centering\let\newline\\\arraybackslash\hspace{0pt}}m{#1}}
\centering
\begin{tabular}{|C{2cm}|C{2cm}|C{2cm}|}
\hline
No. individuals & Annotated frames & Video length (min)\\
\hline
1 & 95,513 & 55\\
2 & 135,547& 119 \\
5 & 44,240& 85\\
10 & 20,321& 91\\
\hline
\end{tabular}
\end{center}
\caption{Dataset Summary: Total number of labeled frames with ground truth data for different group sizes.}
\label{table:datasum}
\end{table}

\subsection{Customization} \label{subsec:Customization}
We release 3D-POPAP (3D-POP Annotation Pipeline) to manipulate the annotations of the dataset (Download: https://github.com/alexhang212/Dataset-3DPOP).
As explained earlier, our use of the 6-DOF tracking decouples the keypoint annotations from the positions of markers used for mo-cap.
Due to this design of the annotation approach, we can offer a unique dataset with the ability to easily add new 2D/3D keypoint annotations.
The feature of keypoint modification is relevant for future work because defining the posture of birds is a difficult problem and depends on the final application.
As of now, there are no datasets available with ground truth on the 3D posture of birds.
The lack of ground truth has motivated novel ideas for solving the 3D reconstruction of bird pose using 2D annotations (silhouette and keypoints \cite{badger_3d_2020}).
Among the available 2D datasets with birds, different numbers of keypoints are selected to define pose \eg CUB-200: 15 \cite{wah_caltech-ucsd_2011}, Cowbird dataset: 12 \cite{badger_3d_2020} and Animal Kingdom: 23 \cite{ng_animal_2022}.

To the best of our knowledge, the use of posture in behavior studies with birds is still limited and pose definition may rely completely on the nature of the study.
Our inspiration for keypoint definition is inspired by gaze studies \cite{kano2022birds,itahara2022corvid} for which 9 keypoint-based posture sufficiently provides gaze direction with body and head orientation.

\subsection{Dataset Validation} \label{ss:Validation}
The annotations in 3D-POP are obtained automatically, and therefore we designed three different tests to validate the accuracy and consistency of the annotations.
The first test compares the accuracy of the 3D features computed with our method and the method presented by Kano \etal \cite{kano2022birds}.

The second test measures the consistency of the 3D/2D annotations across the dataset. 
This test is required to identify errors in annotation introduced by erroneous 3D mo-cap tracking due to occlusion, rapid movement of the birds, or calibration and synchronization errors of the cameras.
It is important to perform this test because manually checking millions of annotations is not practical.
Finally, the third test checks the variation in the 3D pose captured in all sequences.
This test shows that the dataset is not biased to specific types of motion or poses.

\subsubsection{Accuracy} \label{ss:Accuracy}

Kano \etal \cite{kano2022birds} use a calibration method to measure the 3D position of eyes w.r.t. mo-cap markers.
This process involves a custom camera rig, made of 4 separate webcams that capture the head of each pigeon before data collection.
We replicated this process to compute the ground truth 3D position of eyes and beak.
Further, we compared the ground truth with the 3D position of the same features computed with our approach.

We obtained root mean squared errors (RMSE) for all three features (Beak: 5.0 mm, Left eye: 5.0 mm, Right Eye: 4.9 mm), which is sufficient for pigeons considering that the diameter of the eyes is typically 6-7 mm \cite{chard1938structure}. 
This method provides an approximation of the accuracy for a few features only, and a better method is required to test the accuracy of 3D features measured on the body.
It should be noted that our method has comparable accuracy and alleviates the need of using dedicated calibrations rigs and thus saves time.

\begin{table*}
\begin{center}
\newcolumntype{C}[1]{>{\centering\let\newline\\\arraybackslash\hspace{0pt}}m{#1}}
\centering
\begin{tabular}{|p{2.6cm}|C{0.9cm}|C{0.9cm}|C{0.9cm}|C{0.9cm}|C{1.4cm}|C{1.4cm}|C{0.9cm}|C{1.4cm}|C{0.9cm}|}
\hline
RMSE$_{\text{Method}}$ (px) & Beak & Nose & Left Eye & Right Eye & Left Shoulder & Right Shoulder & Top Keel & Bottom Keel & Tail \\
\hline
RMSE$_{\text{BeforeFiltering}}$ & 10.1 & 7.9 & 7.5 & 7.5& 8.4 & 8.7 & 9.4 & 9.9 & 8.8\\
RMSE$_{\text{AfterFiltering}}$ & 8.1 & 6.0 & 5.9 & 5.9& 7.9&8.2&9.1&9.5&8.2\\
RMSE$_{\text{AfterRetraining}}$ & 8.4&6.5 & 6.4 & 6.3 & 8.0 & 8.2 & 9.1 & 9.5 & 8.4\\

\hline

\hline
\end{tabular}
\end{center}
\caption{Root mean squared 2D Euclidean error (px) of each keypoint with different data subsets and trained DLC 2D keypoint models. 
BeforeFiltering: Error of model trained on the full dataset with inference on frames before outliers were filtered. 
AfterFiltering: Errors of the model trained on the full dataset with inference on frames after outliers were filtered. 
AfterRetraining: Errors of the model trained on the filtered dataset with inference on frames after outliers were filtered}
\label{table:Consistency}
\end{table*}

\subsubsection{Consistency and outlier detection}
\begin{figure}
\centering
   \includegraphics[width=0.8\linewidth,trim=0cm 0cm 0cm 0.5cm]{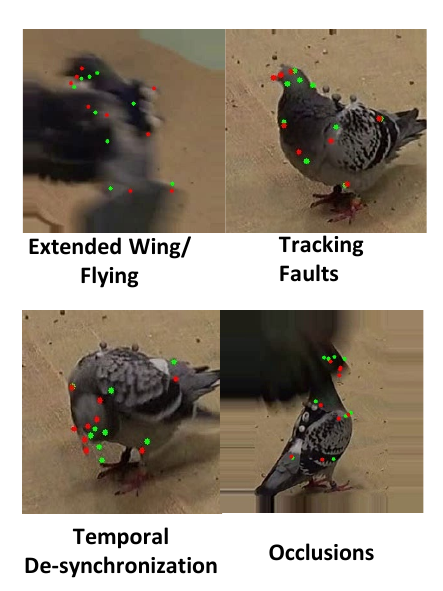}
   \caption{Example frames that are filtered automatically by the outlier analysis, with descriptions of the cause of annotation inaccuracy. Green labels represent annotations, and red labels represent prediction from the trained DLC 2D keypoint detection model.}
\label{fig:FilterExample}
\end{figure}

It is reasonable to assume that a small portion of the mo-cap sequences contains tracking errors and will produce inaccurate 6-DOF poses for body parts.
As a result, the annotation for all keypoints associated with the relevant body parts is likely to be wrong.
We know that models trained with large datasets with small noise still generalize to a solution\cite{rolnick2017deep}.
Yet, it is important to identify and remove these sequences from the dataset. 
Keeping this in mind, we design a consistency check with the intuition that a well-trained model for keypoint detector will predict 2D features with reasonable accuracies for all frames.
Therefore, a comparison between predicted keypoints and propagated keypoints is likely to show very large errors for all keypoints (of the same body part), especially for frames with faulty mo-cap tracking (Figure \ref{fig:FilterExample}).  
We use this idea to automatically determine the consistency of the annotations throughout the trial.

We trained a state-of-the-art 2D keypoint detection model (DLC \cite{mathis_deeplabcut_2018}) on 15177 images with a ResNet50 backbone for 30,000 iterations with the adam optimizer. There is no reliable method available for tracking the posture of multiple birds simultaneously, therefore we use a top-down approach and train on single individual data using bounding box annotations.
The training data excludes highly occluded frames with $>$ 30\% overlap with another bounding box to avoid sequences that have multiple individuals in the bounding box due to close proximity.
GESD outlier analysis \cite{rosner_percentage_1983} is used for each keypoint independently setting the expected outliers at 20\% of the dataset.
The frames having more than 1 outlier keypoint are filtered out as we expect a higher number of outliers in case of erroneous annotations (explained above).

\begin{figure}
\centering
   \includegraphics[width=0.5\textwidth, trim=0cm 2.2cm 2cm 0cm]{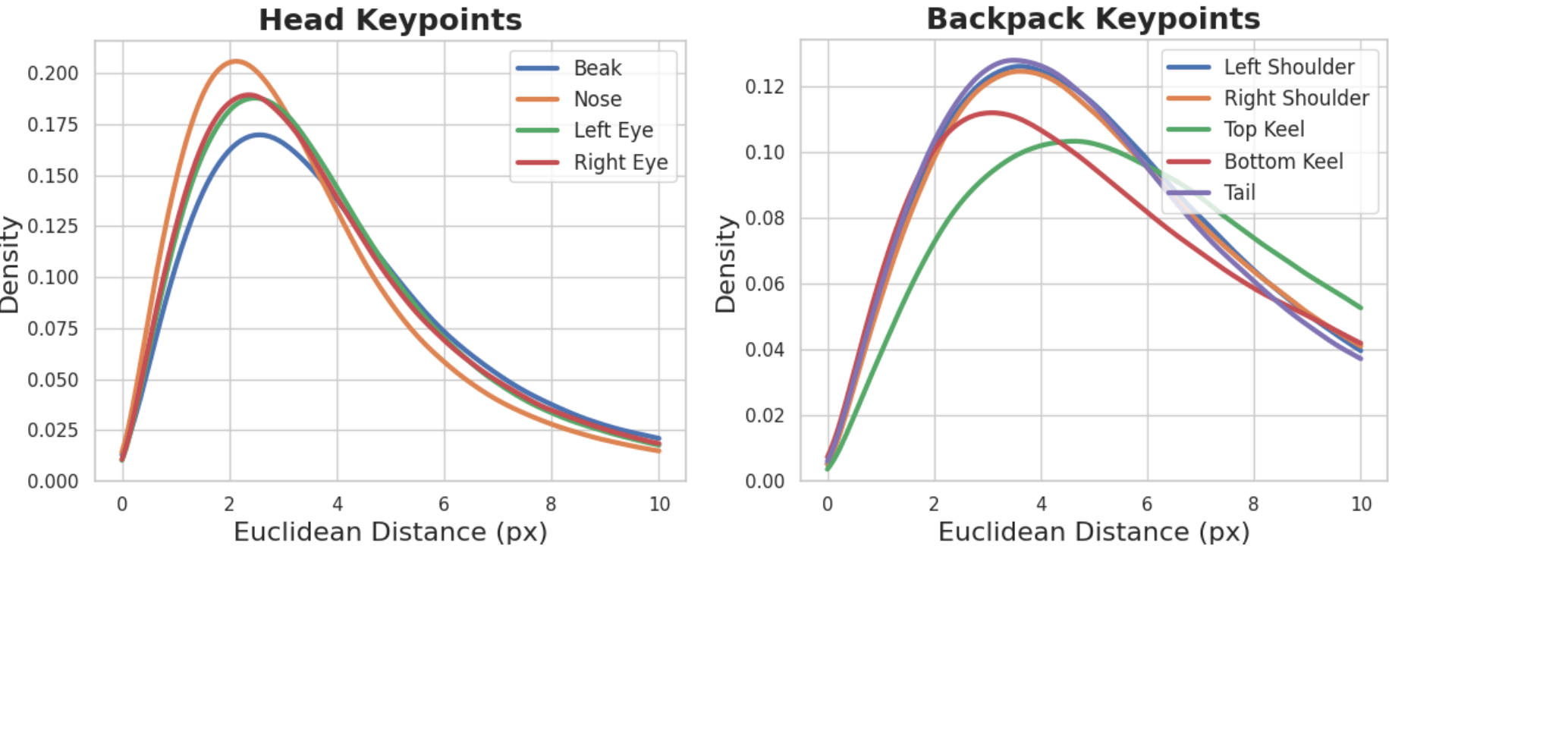}
   \caption{Distribution of Euclidean distances (px) between model predictions of a trained DLC model and annotations, after outlier frames were filtered. Frequency shown in the y-axis and and only points of up to 10px error is shown on the x-axis. A) Head keypoints B) Backpack keypoints}
\label{fig:DLCError}
\end{figure}

Using this method we filtered out 2.9\% of the overall dataset, which lowered the average Euclidean distance between annotation and predictions (see Table \ref{table:Consistency}).
We used the filtered training data and retrained a model (14,722 images, 30,000 iterations, ResNet50 backbone, adam optimizer), but obtained similar errors compared to the previous model (see Table \ref{table:Consistency}). 
The consistency check reveals that the annotations are largely consistent with model predictions, with a typical error of 2-3 pixels for head features and 3-4 px for body features (See Figure \ref{fig:DLCError}).
Figure \ref{fig:FilterExample} shows visual examples of outlier frames where mo-cap errors are likely due to behaviors such as flying or occlusions.  

The outlier filtering method introduces artificial gaps in the dataset. We computed the number of dropped frames and found that 96.1\% of gaps are less than 30 frames (1 second) in length (see supplementary). Researchers in need of continuous temporal data can use gap-free segments or use interpolation to fill small gaps. For sake of completeness, we have included automatically rejected frames in the dataset.

\subsubsection{Pose variation}
We then compute the number of unique poses that each pigeon exhibit to understand the heterogeneity of pose present in the 3D-POP dataset.
It is difficult to compute pose variation directly using the 6-DOF pose defined by markers because the coordinate system is not defined in a standardized way for each pigeon. 
To create a standardized comparison, we compute two planes defined by keypoint features to represent the alignment of the head and body in 3D space.
The head plane is computed using three points (beak and eyes) and the body plane is computed using three points (shoulders and tail). 
In this manner, the orientations of all planes representing the pose of all individuals are defined using the same features and can be compared in a unified coordinate system. 
We use the normal of the planes to compute the angles with the canonical coordinate system (See supplementary). 
It is assumed that a degree of change in rotational angles of either head or body corresponds to a new pose. 
We found a total of 74,924 unique orientations of the head and 14,191 unique orientations of the body, and the combined 1.8 million unique poses present in the dataset. A graphical representation of the range of poses is provided in the supplementary material.

\section{Experiments}

\subsection{Marker-based + Markerless Hybrid Approach} \label{ss:Exp1}
The first experiment shows that markerless tracking algorithm trained on 3D-POP is able to solve 3D tracking for cases when mo-cap fails to track markers.
The solution is useful as an increasing number of pre-existing experimental setups are designed to use marker-based mocap technologies for biological studies \cite{kano2022birds,itahara2022corvid,dunn2021geometric,theunissen_head_2017,kleinheerenbrink2022optimization}.
A hybrid tracking solution, that uses markerless tracking to fill the gaps of the mo-cap system has many potential applications for future behavior studies.

We chose a 5 min sequence with a single individual and artificially removed 25\% of mo-cap tracking data.
The gaps are randomly introduced for a duration of 30-90 frames (1-3 seconds), to mimic tracking loss.
We used the 2D keypoint DLC model (see \ref{ss:Validation}) to detect keypoints from all 4 camera views and triangulate the results with sparse bundle adjustment.
We compared the result with the ground truth and achieved avg. RMS error of 9.2 mm (details in supplementary). 
A simple linear interpolation-based approach to fill gaps resulted in avg. RMS error of 52.1 mm. 
The proposed solution is a viable application because biologists are likely to keep using motion-tracking technology until a robust solution is designed for markerless 3D tracking.
However, we acknowledge that better solutions can be designed for a hybrid approach using temporal consistency in the future \cite{joska2021acinoset}.

\begin{figure}
    \centering
   \includegraphics[width=\linewidth, trim=0cm 1cm 0cm 1cm]{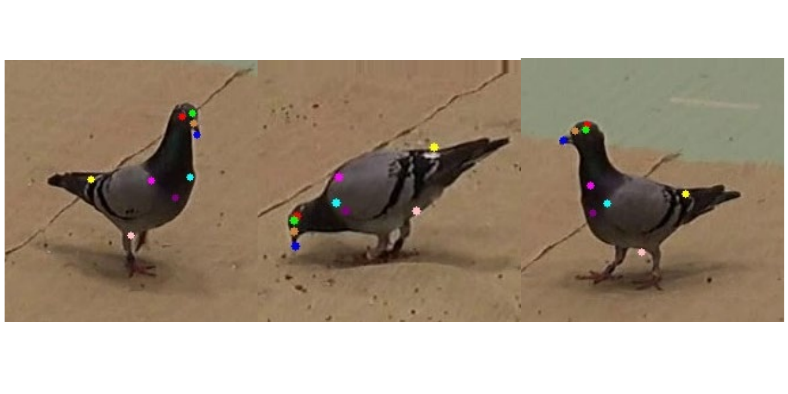}
   \caption{Pictures show that the 2D keypoint detection algorithm trained with the 3D-POP dataset can make predictions on videos with pigeons without any markers attached to the body.}
\label{fig:MarkerlessExample}
\end{figure}
\subsection{Markerless Bird Tracking} \label{ss:Exp2}
This experiment shows that models trained with our datasetcan be directly used to track birds without any markers attached to their bodies.
This experiment works as a ``sanity check'' to ensure that models trained with 3D-POP dataset are not biased toward the presence of markers. 
The test also demonstrates the potential contribution of our method toward developing a complete markerless solution for 3D tracking, posture estimation, and identification. 

Using a pre-trained object detection model (YOLOv5s \cite{redmon2016you}), we extracted the bounding box of a pigeon from a single individual sequence. We then used the 2D keypoint DLC model (see \ref{ss:Validation}) to predict keypoints from the sequence.
The models generalize well to the images of pigeons without markers (see Figure \ref{fig:MarkerlessExample}, supplementary video).
The result is qualitatively checked, but sufficient to prove our claim.
The same solution can be easily extended to multiple pigeon trials by designing a top-down approach (using YOLO) until better solutions are developed using 3D-POP.

\subsection{Manual Validation} \label{ss:Exp3}
This experiment demonstrates the validity of our assumption that keypoints on the body (shoulder, keel, etc.) behave like points on a rigid body. We selected 1000 frames randomly and manually annotated keypoints for the body part. We compared the manual annotations with automatic ground truth annotations using PCK05 and PCK10 (percentage correct keypoint within 5\% and 10\% of bounding box width) metrics. 
We report an average PCK05 of 66\% and PCK10 of 94\% across all keypoints on the body (Table \ref{table:ManualAnnot}).
We also visually quantified that only 2.8\% of the frames are cases where birds are moving their wings, thus the simplified skeletal representation of the body is valid in over 97\% of the dataset.

\begin{table}[h]
\begin{center}
\newcolumntype{C}[1]{>{\centering\let\newline\\\arraybackslash\hspace{0pt}}m{#1}}
\begin{tabular}{|p{1cm}|C{1.1cm}|C{1.1cm}|C{0.7cm}|C{1.1cm}|C{0.7cm}|}
\hline
Metric & Left Shoulder & Right Shoulder & Top Keel & Bottom Keel & Tail \\
\hline
PCK05 & 0.78 & 0.75 & 0.58 & 0.57&0.60\\
PCK10 & 0.98 & 0.98 &0.94&0.89&0.92\\
\hline
\end{tabular}
\end{center}
\caption{PCK errors per body keypoint between manual annotation and 3DPOP annotation. PCK is defined as the percentage of points that are within 5\% and 10\% of the bounding box width}
\label{table:ManualAnnot}
\end{table}
  
\section{Limitations and Future work }
The annotation method presented in the paper largely relies on the assumption that the head and body mostly behave as rigid bodies. 
This assumption does not hold for certain body parts such as the neck, tail end, or feet and limits the selection of keypoints at these body parts.
For similar reasons, the proposed approach will not support annotation for flying birds or birds that change the shape of body parts while performing certain behaviors \eg courtship \cite{janisch_deciphering_2021}.

Our approach inherently depends on the tracking accuracy of the mo-cap system.
Users must maintain mo-cap systems regularly calibrated for consistent results.
Another possible source of error in the annotation pipeline is video camera calibration and its temporal synchronization with the mo-cap system.
We do show that our outlier detection method is effective at identifying noisy annotations, however, noise can still be present in the dataset. 
Finally, since the dataset was curated semi-automatically in an existing motion tracking setup, the data we provide is limited to an indoor environment. 

We have improved the existing state of the art for multi-animal tracking by adding complexity in the form of the number of individuals and camera views.
In the future, we intend to develop lifting-based approaches \cite{gunel2019deepfly3d,gosztolai_liftpose3d_2021} to learn the 2D-3D mapping obtained in the 3D-POP dataset to track birds in outdoor environments.

\section{Conclusion}
In this paper, we introduced a novel method to use a mo-cap system for generating large-scale datasets with multiple animals. 
We demonstrate that our semi-automated method offers an alternative for generating high-quality datasets with animals without manual effort. 
We offer 3D-POP, the first dataset with ground truth for 3D posture prediction and identity tracking in birds, which is extremely difficult to achieve even with manual labor.
3D-POP dataset offers an opportunity for the vision community to work on a complex set of vision problems relevant to achieving markerless tracking of birds in indoor and outdoor environments. 
At the same time, our method will motivate biologists to create new datasets as they have access to and work with different types of animals. 

\section{Acknowledgements}
Funded by the Deutsche Forschungsgemeinschaft (DFG, German Research Foundation) under Germany’s Excellence Strategy – EXC 2117 (ID: 422037984). 
The Ethical Committee of Baden-Württemberg approved all the experiments(Regierungspräsidium Freiburg, Referat 35, License Number: 35-9185.81/G-19/107).
M.N. acknowledges additional support from the Hungarian Academy of Sciences (grant no. 95152) and Eötvös Loránd University.
I.C. also acknowledge Office of Naval Research (grant ONR, N00014-19-1-2556), Horizon Europe Marie Sklodowska-Curie Actions (860949) and the Max Planck Society.
\newpage
{\small
\bibliographystyle{ieee_fullname}
\bibliography{egbib}
}

\end{document}


\maketitle

\begin{abstract}
    The following text is supplementary text for the paper "3D-POP - An automated annotation approach to facilitate markerless 2D-3D tracking of freely moving birds with marker-based motion capture". 
    The text includes details of methods and results that are not part of the main text which are described in more detail here. We also outline detailed method that can be helpful to replicate the setup and annotation process, especially for biologists. 
\end{abstract}


\section{Affiliations}
Here, we provide complete affiliations for the authors from the main text, with identical numbering.

1. Department of Collective Behaviour and Department of Ecology of Animal Societies, Max Planck Institute of Animal Behavior, 78464 Konstanz, Germany.

2. Department of Biology, University of Konstanz, 78464 Konstanz, Germany.

3. Centre for the Advanced Study of Collective Behaviour, University of Konstanz, 78464 Konstanz, Germany.

4. Computer Aided Medial Procedures, Informatik Department, Technische Universität München, Boltzmannstraße 3, 85748, Garching bei München, Germany. 

5. Department of Biological Physics, Eötvös Loránd University, Pázmány Péter sétány 1A, Budapest 1117, Hungary.

6. MTA-ELTE ‘Lendület’ Collective Behaviour Research Group, Hungarian Academy of Sciences, Budapest 1117, Hungary.

\section{Supplementary Methods}
\subsection{Camera Calibration:}
The Infrared cameras of the vicon motion capture system (Vicon Vero,Vantage) are calibrated using built in software (Vicon Nexus) with a calibration wand. 
All cameras are also time synchronized when recording. 
The calibration of the Vicon system forms the basis of the whole dataset, in which we use the vicon coordinate system for all 3D coordinates. 

For the 4 high definition RGB action cameras, we performed intrinsic calibration, extrinsic calibration and time synchronization independently. 

\subsubsection{Intrinsic and Extrinsic Calibration}
For intrinsic calibration, we used an A0 (84.1 x 118.9 cm) charuco checkerboard before each day of recordings and undistorted all videos from each camera view using the obtained distortion and camera matrix from opencv.

For extrinsic calibration, we adopted a subject based approach, where we manually annotated the 2D position of a motion capture markers visible in the image (\eg, backpack marker) on a moving pigeon subject over up to 30 frames. 
For each frame, we compute camera pose using 2D marker positions on the backpack and the 3D coordinates of the backpack in the vicon coordinate system.
The combination of both provide us the extrinsic parameters for each camera. 
We ensure that sampled 3D positions are well distributed in the tracking volume to avoid bias in extrinsic parameters. 
This approach is useful as it allows us to move the tripod positions between sessions and perform fast extrinsic calibration without using the checkerboard. 

In the future, we plan to make the method for marker selection automatic, which would improve accuracy as the system will recompute extrinsics in real-time and change in camera position would not require calibration.


\subsection{Temporal synchronization}
To synchronize the RGB action cameras, we attached a camera control box (Sony CBB-WD1) to each action camera. The control boxes has built-in functionality to synchronize video streams from multiple cameras over ethernet and a network switch. 

Since our data are collected by two independent systems (Motion tracking system and RGB cameras), we designed an arduino based synchronization device with 3 RGB LED lights and 2 infra-red lights that blinks for 1 second at 5 second intervals. For the RGB videos, we computed the change in maximum pixel values of the cropped box area through time, and detected light flashes based on a change of more than 30 units. For the vicon system, we attached 4 additional markers onto the arduino box, which allow a new object to be defined within the mo-cap software together with the 2 infra-red lights. Flashes can then be detected based on the number of markers present in the object (6 markers for flash on, 4 markers for flash off). The systems are then temporally synchronized by matching the detected light flashes from both systems. In cases where a light flash were not detected, we manually filled in the flash by assuming the flash is exactly 6 seconds after the previous one. 

\subsection{Markerless Data}
POP-3D dataset contains ground truth annotation for pigeons with markers attached to their body. However, we expect this dataset to play a role in development of markeless algorithms for tracking pigeons and other birds. 
In experiment 2, we show that models trained on 3D-POP dataset do work well with pigeons recorded in the same arena without any markers. 
To validate results in future methods, 3D-POP dataset also includes trials of freely moving birds without any marker attachment (n = 1,2,5,11). 
The videos will serve useful for making qualitative claim about performance of algorithms developed with mocap annotated posture or identity data. 
It is worth noting that we only demonstrate that position of markers do not play role in prediction of keypoints but do not show same validation for problem of identity recognition. 
Markers may play a role in identity recognition, we would like to test it in future work. 
Please see Table \ref{table:datasum-womarkers} for more details of sessions recorded with pigeons without markers.

\begin{table}[h]
\begin{center}
\begin{tabular}{|p{2cm}|p{2cm}|p{2cm}|}
\hline
No. individuals & Available frames & Video length (min)\\
\hline
1 & 36,825 & 20\\
2 & 36,600 & 20 \\
5 & 9,810& 5.5\\
11 & 36,225& 20\\
\hline
\end{tabular}
\end{center}
\caption{Markerless Data Summary: Total number of frames and video length for markerless data.}
\label{table:datasum-womarkers}
\end{table}

\subsection{Post-processing Mo-cap data}
The data obtained from motion capture is often not directly usable for the annotation pipeline. 
It requires a post-processing step to ensure smooth annotation process. 
Marker-based motion capture has a limitations related to tracking loss and marker identification (for 6-DOF tracking).
This results in error of identification of pigeons and 6-DOF pose of rigid bodies defined by the mo-cap system. 
The problem stems from limited availability of space on the pigeon body. 
The markers are forced to be close to each other (in 4-marker pattern) which leads to error in correspondence matching required for pose computation. 
To solve for these changes, we applied a post-processing pipeline introduced by Kano \etal \cite{kano2022birds} to fix mis-labelled frames by detecting large changes in the distance between defined markers, then determining the correct labels through permutation techniques.
Detailed description and code used is provided in Kano \etal \cite{kano2022birds}.

\begin{figure*} [!h]
   \includegraphics[width=\textwidth,trim=0cm 1cm 3cm 0cm]{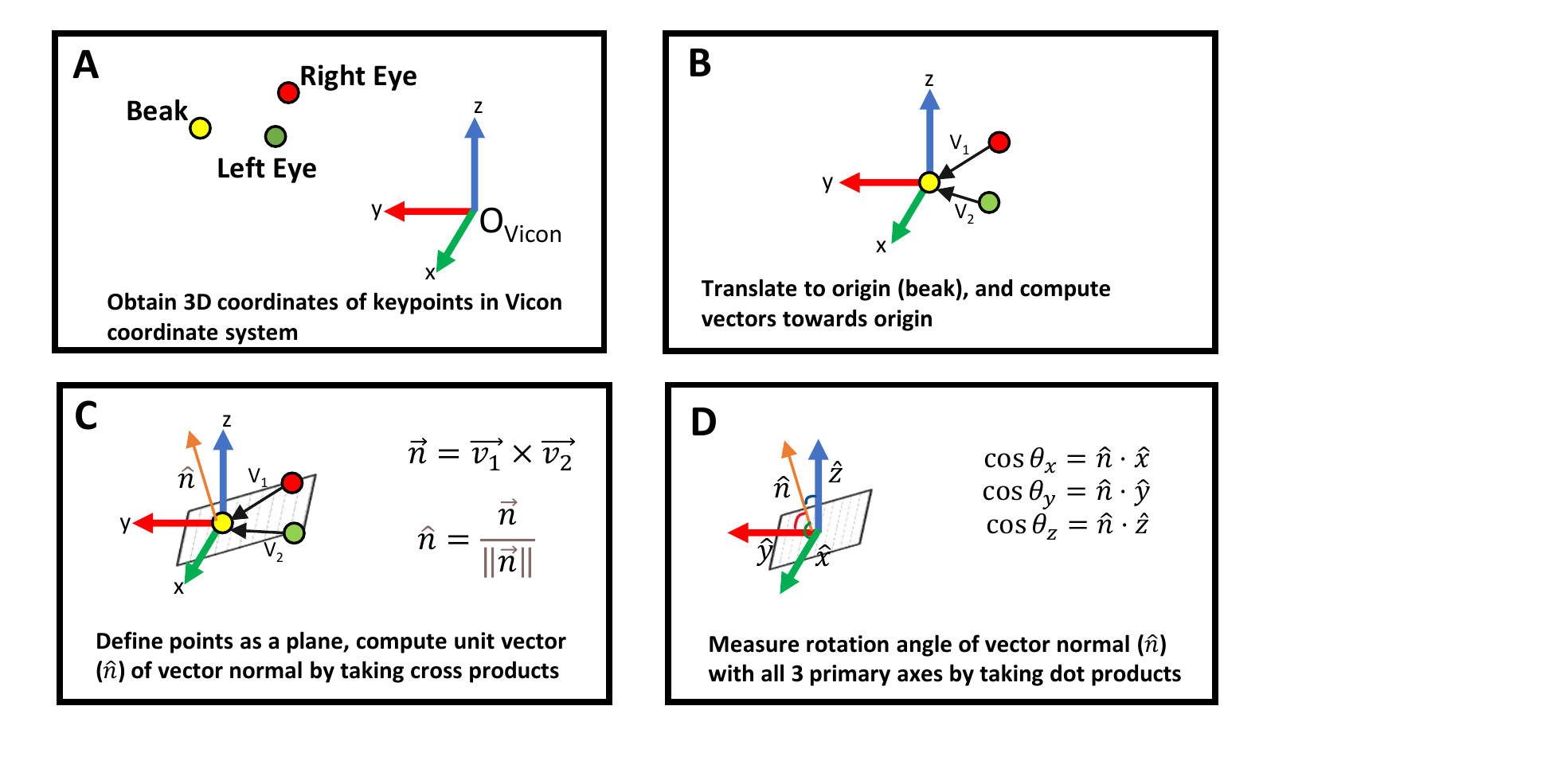}
   \caption{Schematic for computing 3D orientation angles of the pigeon's body parts (head and body) for a given frame. A) Obtain 3D coordinates of keypoints in the world coordinate system (vicon). B) Translate all keypoints to new origin (beak) and compute vectors towards the origin using keypoints. C) Define a plane and compute surface normal. D) Calculate angle between surface normals and the 3 primary axes.}
\label{fig:computePose}
\end{figure*}

\begin{figure*}
   \includegraphics[width=\textwidth,trim=0cm 1cm 1cm 0cm]{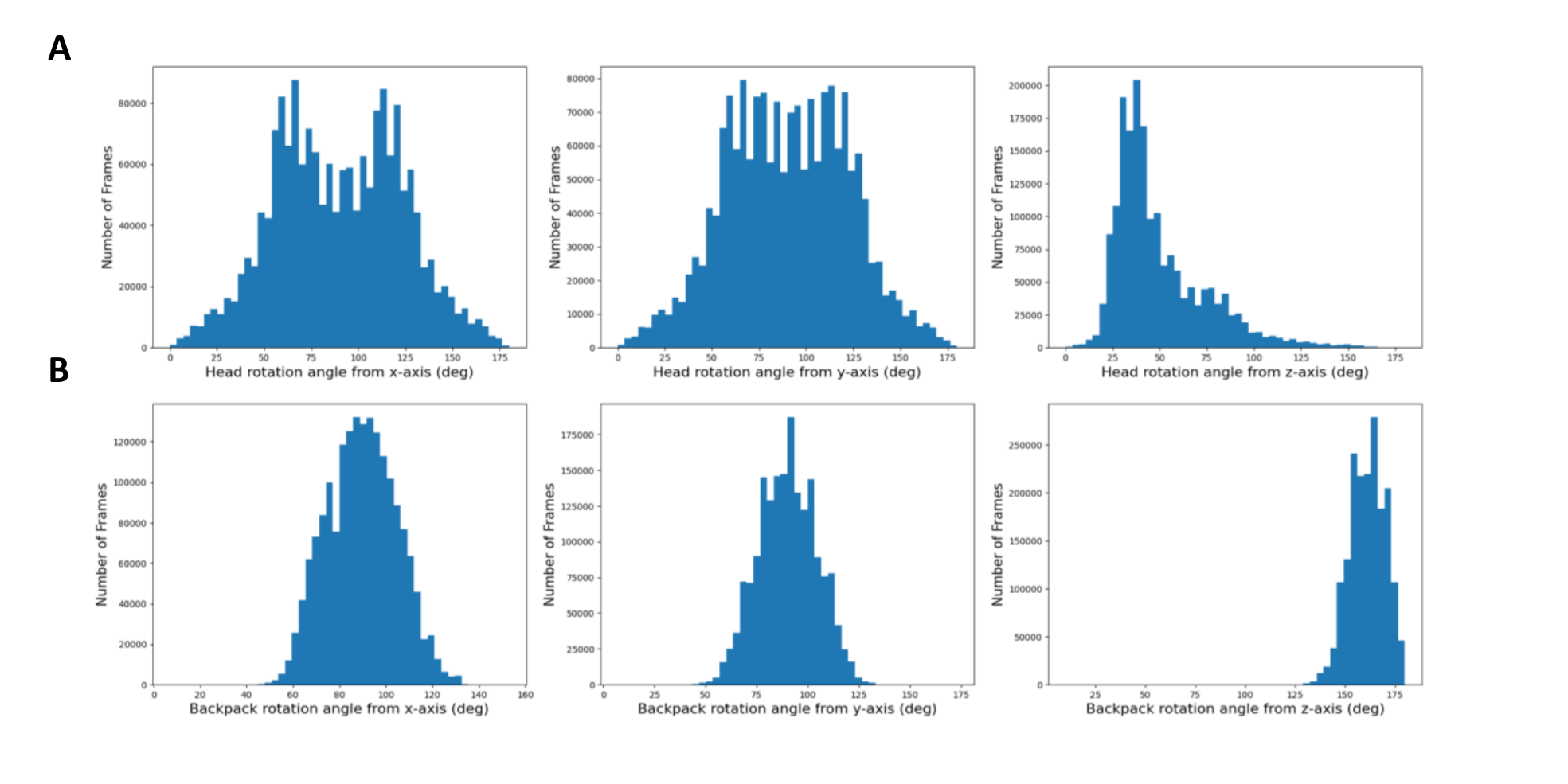}
   \caption{Frame distribution of the head and backpack rotation angles with respect to the 3 primary axes of pigeon subjects present in the dataset. A) Distribution of head rotation angles. B) distribution of backpack rotation angles. }
\label{fig:PoseHistogram}
\end{figure*}

\subsection{Computing Pose Variation:}
Computing variation in pose is important to understand how many different postures are recorded in the dataset. 
Definition of coordinate system defined for each pigeon's body part (head, body) is slightly different because the definition is based on marker positions, which is different for each pigeon each day.
Therefore, the orientation angles  defined by the 6-DOF pose w.r.t the canonical frame (world coordinate system) are different for each pigeon even if absolute posture of pigeons is the same.
This problem does not allow us to compute variation of pose for each pigeon in a standard manner.
We argue that pigeon posture can be measured in a standard way if orientation of the body parts are defined using positions of 3D keypoints in a unified coordinate system \ie, world coordinate system. 
The keypoints such as beak, eyes or shoulders are common features recorded for each bird and thus a posture representation involving these features would provide means for comparing postures of different pigeons.
Using this logic we designed a new technique to measure 3D orientation (rotational angles)  of pigeon body parts relative to each axis. 
Please refer Figure \ref{fig:computePose} for a pictorial representation. 
 
In this text, we will describe the steps to obtain rotation angles of the head in the world coordinate system. 
Firstly, we select beak as primary keypoint and shift the origin of world coordinate system to this point (translation).
This is done because we are interested in comparing rotation units and shifting origin reduces complexity of pose representation from 6-DOF (rotation and translation) to 3-DOF (rotation).
In other words, we change the representation of a rigid body (pigeon head) from 6-DOF pose representation to a plane representation (passing through origin).
The plane is defined using 3D positions of beak, left eye and right eye (Figure \ref{fig:computePose}). 
The normal of the plane is the cross product of two vectors originating from beak to left and right eye.
This normal is defined at the origin and it's angles with respect to the primary x-y-z axis, which represent one unique head orientation in the world coordinate system.
This process is repeated to compute head posture of all pigeons in all sequences.
The comparison between all posture is only possible because the world coordinate system is consistent for all sequences. 
Finally, we show a histogram of the occurrences of the different rotation angles to indicate pose variation  (Figure \ref{fig:PoseHistogram}).
The same process is repeated for body pose by defining a plane using shoulder and tail keypoints, with tail as origin.

\subsection{Dataset comparison}
There are many datasets available with animals that target one or more problems. 
We provide a short overview 18 different datasets and provide a table at the end of this text as auxiliary information. 

\section{Experimental results}

\subsection{Outlier Detection and Filtering Pipeline}
Our outlier detection and filtering pipeline introduces gaps in the dataset. Here is a quantification of the number and length of gaps that are present in the dataset. (Table \ref{table:Gaps})

\begin{table}[h]
\begin{center}
\begin{tabular}{|p{2cm}|p{2cm}|p{2cm}|}
\hline
1 frame & 2-30 frames & $>$30 frames\\
\hline
3585 & 5062 & 351\\
\hline
\end{tabular}
\end{center}
\caption{Frame length of consecutive gaps present in the dataset}
\label{table:Gaps}
\end{table}


\subsection{Experiment 1 - Hybrid Approach:}
In the paper, we performed an experiment to show that markerless tracking is possible for pigeons using the 3D-POP dataset. 
First, experiment was performed with pigeons with marker attached, like experimental scenario of Kano \etal \cite{kano2022birds}.
We claim that markerless solution is directly useful to improve tracking performance for cases where motion capture is already in use.
We took a sample data from our recording sequence and introduced gaps in trajectories to simulate loss of tracking, then further used markerless approach to fill the gaps to evaluate quality of markerless 3D tracking in comparison with ground truth.
The results are demonstrated in Table \ref{table:Hybrid}, which show that using a 2D keypoint detection model and simple triangulation, the gap filling algorithm provides good accuracy.
This is useful already for experiments where missing data with mo-cap setup is a consistent problem. 
We also show a simple comparison with interpolation approach to show that markerless solution have higher chance of filling gaps than data interpolation methods. 
In future work, we want to try different strategies to combine multi-view data to get higher prediction accuracy.

\begin{table*}[ht]
\begin{center}
\begin{tabular}{|p{2cm}|p{1cm}|p{1cm}|p{1cm}|p{1cm}|p{1cm}|p{1cm}|p{1cm}|p{1cm}|p{1cm}|}
\hline
RMSE$_{\text{Method}}$ \newline (mm) & Beak & Nose & Left Eye & Right Eye& Left Shoulder & Right Shoulder & Top Keel & Bottom Keel & Tail \\
\hline
RMSE$_{\text{Hybrid}}$ & 8.2&6.5&7.3&6.3&13.8&9.4&13.3&8.9&9.2\\ 
RMSE$_{\text{Linear}}$ & 66.3 & 64.8 & 62.5& 62.7 & 45.5 & 42.7 & 41.3 & 37.6 & 45.7\\
\hline
\end{tabular}
\end{center}
\caption{Root mean squared Euclidean error (mm) of different approaches used to fill artificially introduced gaps in a 5 min single pigeon sequence. Hybrid approach uses a 2D DLC model from each view and triangulated and the linear approach interpolates missing data linearly with data before and after a given gap.
}
\label{table:Hybrid}
\end{table*}

\section{Additional Files}
We have provided additional material along with this file. 
\begin{itemize}
    \item Supplementary video : The video is designed to introduce the reader with 3D-POP dataset. It shows the setup, diversity of the dataset and the annotations on video images. 
    
    Click \href{https://youtu.be/er4u0WpRJeQ}{here} for youtube link of the video. 
    

    
\end{itemize}


\bibliographystyle{plain} 
\bibliography{egbib}